\documentclass[runningheads]{llncs}

\usepackage[finail,year=2024,ID=10448]{eccv}
\usepackage{tabularray}

\usepackage{eccvabbrv}

\usepackage{graphicx}
\usepackage{booktabs}
\usepackage{multirow}
\usepackage{algorithm}
\usepackage{algpseudocode}
\usepackage{multicol}
\usepackage{tabularx}

\usepackage[accsupp]{axessibility}  

\usepackage[pagebackref,breaklinks,colorlinks,citecolor=eccvblue]{hyperref}

\usepackage{orcidlink}

\begin{document}

\title{Refining Text-to-Image Generation: Towards Accurate Training-Free Glyph-Enhanced Image Generation}
\titlerunning{SA-OcrPaint}





\author{Sanyam Lakhanpal \inst{1} \and
Shivang Chopra  \inst{2} \and
Vinija Jain\inst{3} \and \\
Aman Chadha\inst{3,4}\thanks{Work does not relate to position at Amazon.} \and 
Man Luo\inst{1}
} 

\authorrunning{S.~Lakhanpal et al.}

\institute{Arizona State University\\
\email{\{slakhanp,mluo26\}@asu.edu}
\and Georgia Institute of Technology\\ 
\email{shivang-chopra11@gatech.edu}
\and Stanford University \and Amazon AI\\
\email{hi@vinija.ai, hi@aman.ai}
}

\maketitle

\begin{abstract}
Over the past few years, Text-to-Image (T2I) generation approaches based on diffusion models have gained significant attention. However, vanilla diffusion models often suffer from spelling inaccuracies in the text displayed within the generated images. The capability to generate visual text is crucial, offering both academic interest and a wide range of practical applications. To produce accurate visual text images, state-of-the-art techniques adopt a glyph-controlled image generation approach, consisting of a text layout generator followed by an image generator that is conditioned on the generated text layout. Nevertheless, our study reveals that these models still face three primary challenges, prompting us to develop a testbed to facilitate future research.
We introduce a benchmark, LenCom-Eval, specifically designed for testing models' capability in generating images with \textbf{Len}gthy and \textbf{Com}plex visual text. 
Subsequently, we introduce a training-free framework to enhance the two-stage generation approaches. We examine the effectiveness of our approach on both LenCom-Eval and MARIO-Eval benchmarks and demonstrate notable improvements across a range of evaluation metrics, including CLIPScore, OCR precision, recall, F1 score, accuracy, and edit distance scores. 
For instance, our proposed framework improves the backbone model, TextDiffuser,  by more than 23\% and 13.5\% in terms of OCR word F1 on LenCom-Eval and MARIO-Eval, respectively.
Our work makes a unique contribution to the field by focusing on generating images with long and rare text sequences, a niche previously unexplored by existing literature\footnote{Upon acceptance of the paper, both the datasets and the code will be made public.}. 

\end{abstract}

\section{Introduction}
\label{sec:intro}

In recent years, the field of computer vision (CV) has seen remarkable advancements, particularly in the domain of image-generation models conditioned on textual descriptions~\cite{balaji2022ediffi,ho2020denoising,song2020denoising,gal2022image,gu2022vector,rombach2022high,saharia2022photorealistic,zhang2023adding,zhao2024uni}. These models, leveraging sophisticated Generative Adversarial Networks (GANs)~\cite{goodfellow2014generative}, Transformer model~\cite{vaswani2017attention} and Denoising Diffusion Probabilistic Models~\cite{sohl2015deep,ho2020denoising}, have demonstrated an extraordinary capability to translate textual prompts into vivid, detailed images, opening new frontiers in digital art design, user interaction, material design and medical image reconstruction~\cite{yang2023diffusion}. Despite their impressive achievements, these models exhibit a notable limitation when it comes to generating images that include specific visual text~\cite{liu2023character,chen2023textdiffuser,yang2024glyphcontrol,chen2024textdiffuser,tuo2023anytext}.

Visual text image generation is a critical task with wide-ranging applications, touching everything from advertising, where the precise rendering of brand names on products can alter consumer perception, to educational resources, where accurate depictions of text in diagrams or illustrations can significantly impact learning outcomes. 
However, the state-of-the-art diffusion models often struggle with this task~\cite{rombach2022high}. 
To enhance the visual text performance, a series of studies have deployed a two-stage diffusion pipeline~\cite{chen2023textdiffuser,chen2024textdiffuser,yang2024glyphcontrol,tuo2023anytext}, where the first stage generates glyph images, rendering keywords\footnote{In this work, keywords refer to the words indented to be generated in an image.} with specific fonts and sizes on a whiteboard, and the second stage involves glyph-controlled image generation. 
Although these models indeed improve the fidelity of generated visual text, they generally fall short in rendering longer textual elements.
This limitation not only restricts the utility of these models but also highlights a gap in our understanding of how to effectively integrate textual and visual information in generated content.

Moreover, the precise generation of the text specified in the input prompt is of utmost importance. Users may require the inclusion of specific terms, names, or phrases—often unique or novel—in the generated images. These could range from the name of a new coffee blend on a menu to an innovative company logo. Existing models, however, tend to default to generating more common words or phrases, substituting the requested unique terms with generic alternatives. This not only detracts from the personalized experience sought by users but also diminishes the model's utility in applications requiring a high degree of specificity and accuracy in text representation.

\begin{figure}[t]
    \centering
    \includegraphics[width=0.98\linewidth]{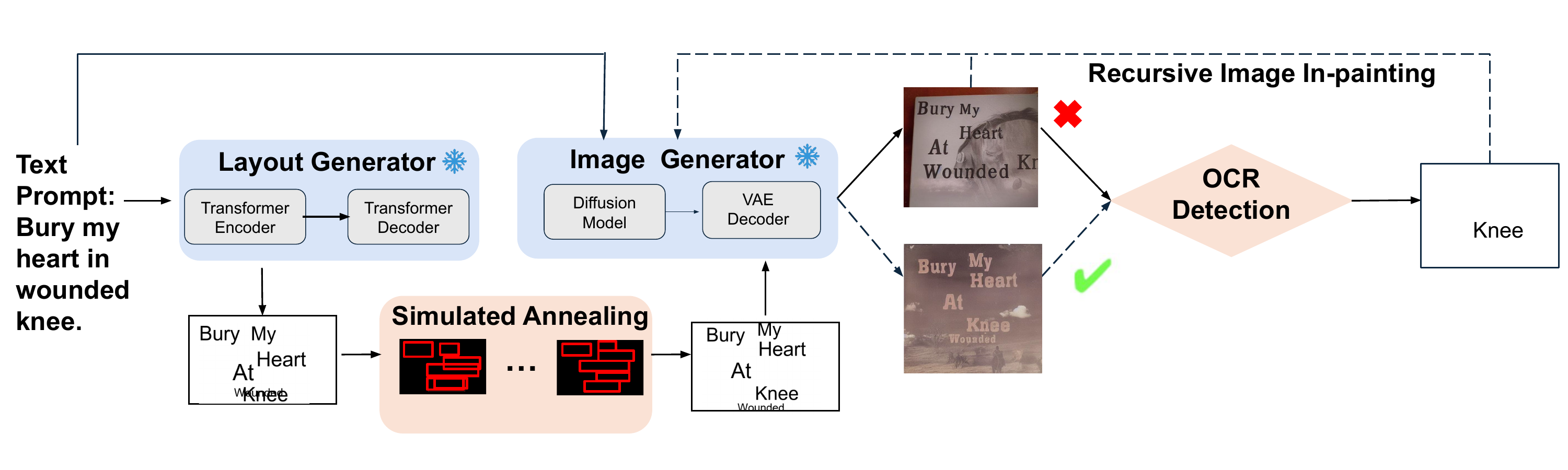}
    \caption{The proposed training-free framework to improve overall accuracy of visual text generation. This method consists of two main stages. First, it minimizes the overlapping of keyword bounding boxes created by the layout generator. Following that, OCR is employed to identify any spelling errors, following which we give a new mask region and the generated image to the pretrained in-painting image generation model. The second step is applied recursively.}
    \label{fig:framework}
\end{figure}

Our research investigates the intricate challenge of generating images with embedded long sequences of visual text. We first developed a specialized evaluation benchmark, named for LenCom-EVAL, to evaluate a model's ability to generate \textbf{Len}gthy and \textbf{Com}plex visual text images.
Through rigorous analysis of current image generation models using our dataset, we uncovered three major weaknesses. These include diminished performance with increased text length, poor layout generation leading to overlapping text, and the models' inability to strictly adhere to text prompts. 
In response to these findings, we propose a training-free framework, as illustrated in Figure~\ref{fig:framework}, aimed at enhancing both the text layout and image generation processes to address the identified limitations.

Our method initially employs a simulated annealing algorithm~\cite{kirkpatrick1983optimization} to minimize overlap between rendered keywords in the initial glyph image. Despite being simple, this step substantially improve the glyph-controlled image generation.
Nevertheless, the generated images still contain misspellings, such as missing characters, extra characters, or merged characters. To further enhance the overall visual text accuracy, we design an OCR-Aware recursive in-painting technique. This involves initially using OCR to detect any misspellings in the generated glyph image, subsequently generating a mask feature map and utilizing an off-the-shelf image in-painting method to correct these misspellings in the initially generated image. This process is applied recursively to ensure the generation of visually accurate text images. 
Our framework, named SA-OcrPaint, stands for \textbf{S}imulated \textbf{A}nnealing and \textbf{OCR}-Aware Recursive In-\textbf{Paint}ing for Glyph-Enhanced Image Generation.
Notably, SA-OcrPaint does not require any training process and can leverage the capabilities of pretrained diffusion models. It can be seamlessly integrated with any two-stage glyph-controlled image generation models. 
In this work, we showcase the proposed framework's effectiveness in conjunction with TextDiffuser~\cite{chen2024textdiffuser}, on both 
LenCom-EVAL and MARIO-EVAL~\cite{chen2024textdiffuser} via multiple metrics. 
For instance, SA-OcrPaint achieves 23\% and 13.5\% improvement in terms of OCR word-level F1 on the two benchmarks respectively. 
SA-OcrPaint set the new state-of-the-art results on MARIO-EVAL.

In summary, our contributions are threefold:

(i) We address the underexplored challenge of generating images with lengthy visual text, introducing a benchmark (LenCom-EVAL) that provides valuable insights into this complex area. This effort is crucial for advancing the design and application of generative models.

(ii) By identifying the shortcomings of existing models in accurately generating images with visual text, our research illuminates the areas needing improvement, such as enhancing text-image coherence, refining layout generation, and ensuring models can precisely follow text prompts.

(iii) We propose an innovative training-free method to enhance glyph-controlled image generation model, enabling the creation of more accurate images with embedded text, marking a significant advancement in the field of visual text image generation.

\section{Related Work}

\paragraph{Visual Text Image Generation Datasets.}
There are some general and widely used image-pairs datasets~\cite{lin2014microsoft,gu2022wukong,schuhmann2022laion,schuhmann2021laion}. However, these datasets are not tailored for the purposes of text rendering or the assessment of spelling accuracy.
On the other hand, there are some work aware of the importance of visual text generation and provide benchmark along this research. 
The SynthText in the Wild dataset \cite{gupta2016synthetic} offers a synthetic compilation where word instances are integrated into images of natural scenes. Meanwhile, TextSeg \cite{xu2021rethinking} provides a collection of text images sourced from everyday environments, including but not limited to greeting cards and signage found on roads.
The DrawText benchmark \cite{saharia2022photorealistic} is divided into two segments. The first, DrawText Spelling, leverages a templated approach to generate text by filling placeholders with sampled words. The second part, DrawText-Creative, features 175 unique prompts that challenge the rendering of text across a spectrum of creative styles and contexts, which were conceived by a seasoned graphic designer.
MARIO-10M dataset \cite{chen2024textdiffuser} encompasses 10 million image-text pairs. These pairs are annotated for text recognition, detection, and character-level segmentation, sourced from LAION-400M \cite{schuhmann2021laion}, The Movie Database (TMDB)\footnote{https://www.themoviedb.org}, and Open Library. Additionally, MARIO-EVAL, a dataset formulated exclusively for evaluation, comprises a selection from MARIO-10M, DrawBenchText \cite{saharia2022photorealistic}, DrawTextCreative \cite{liu2022character}, and ChineseDrawText \cite{ma2023glyphdraw}.
AnyWord-3M~\cite{tuo2023anytext} is designed for multilingual evaluation, which encompasses 3 million pairs of images and texts with OCR annotations.
While preceding studies have primarily concentrated on either brief text segments or the exploration of multilingual content and creativity, our investigation sets itself apart by focusing on the generation of lengthy and uncommon visual text.

\paragraph{Visual Text Image Generation Models.}
While Stable Diffusion (SD) models excel at producing visually compelling images, they struggle to generate coherent and accurate visual text~\cite{ramesh2022hierarchical}. To address this limitation, recent advancements have focused on enhancing SD models with character-aware modules \cite{liu2023character} and text layout generation modules \cite{chen2024textdiffuser,yang2024glyphcontrol,chen2023textdiffuser}, significantly improving the fidelity of visual text image generation. 
The introduction of a Character-aware text encoder \cite{liu2023character}, demonstrates a reduction in spelling errors compared to models utilizing Character-blind encoders. 
UDiffText \cite{zhao2023udifftext} further innovates by training a character-aware text encoder alongside an inpainting version of stable diffusion \cite{rombach2022high}, curated from extensive datasets. This approach highlights the inadequacy of standard denoising loss for text rendering tasks, leading to the introduction of a local attention map to better capture character regions. 
Additionally, a noise-adjusted latent refinement process is applied during inference to minimize spelling inaccuracies.
TextDiffuser~\cite{chen2024textdiffuser}, integrates a two-module system wherein the layout generation module generates the textual layout (derived from the input prompt), which is subsequently processed by an image generation model derived from U-Net \cite{ronneberger2015u}. This system employs a modified layout transformer \cite{gupta2021layouttransformer} for text rendering, while TextDiffuser-2 \cite{chen2023textdiffuser} leverages a language model for more varied layout planning. Enhancements in spelling accuracy are achieved through the incorporation of fine-grained tokenizers, such as character and position tokens, inspired by \cite{liu2023character}.
Diff-Text \cite{zhang2023brush} presents a training-free framework aimed at generating multilingual visual text images. It utilizes localized attention constraints and contrastive image-level prompts within the U-Net's cross-attention layer to improve the accuracy of textual region.
AnyText~\cite{tuo2023anytext} can generate multilingual text in the image by introducing a novel text embedding module and a text perceptual loss. 
Our research identifies limitations in existing methodologies and devising training-free techniques to refine TextDiffuser for the generation of lengthy and uncommon visual text, thereby pushing the boundaries of visual text image generation.

\section{Lengthy and Complex Visual Text Evaluation Datasets}

Visual text generation holds significant potential across various aspects of daily life.
However, existing datasets such as MARIO-EVAL~\cite{chen2024textdiffuser} for visual text generation remain in a nascent phase, primarily focusing on the generation of short textual content or emphasizing the font and diversity of text, rather than assessing comprehensive system capabilities. In contrast, our goal is to gauge the model's efficacy in processing lengthy textual elements but also to challenge it with rare or intentionally complex scenarios. Addressing lengthy and uncommon visual text situations is crucial, as it ensures the model's robustness and adaptability, preparing it for a wider range of real-world applications where unpredictability and complexity are commonplace. 
To this end, we have developed a testbed, named as LenCom-EVAL,  that includes three subsets described in the following. Table~\ref{tab:statistic} shows the statistics for each subset, compares them to those of MARIO-EVAL, and demonstrates that our datasets serve as a complementary evaluation set.
We provide the word distribution of LenCom-EVAL and the comparison with MARIO-EVAL in Appendix~\ref{apd:dataset_distribution}. 
All three subsets only include the input text prompts since we focus on the accuracy of the generated text (cf. \S\ref{sec:eval_metrics} for a discussion on evaluation metrics). 
Example from each subset can be found in Appendix~\ref{apd:dataset_examples}. 

\paragraph{MARIO-Hard.} 
We selected a subset from MARIO-EVAL \cite{chen2024textdiffuser}, originally created from the LAION dataset. The selection criterion was that each example must contain four or more keywords in the generated image. This subset is considered to consist of hard samples from MARIO-EVAL and serves our objective of evaluating the model's efficiency in processing and generating images with lengthy textual content.

\paragraph{Aug-MARIO-Hard.} 
We utilize three augmentation strategies identified as \textit{spelling}, \textit{keyboard}, and \textit{splitting}, where \textit{spelling} introduces spelling errors into a word, \textit{keyboard} modifies a word by incorporating adjacent keyboard letters, and \textit{splitting} divides a word into two segments. 
The motivation behind augmentation is our desire to infuse the dataset with attacked words, thereby testing the model's fidelity to the text prompts and evaluating its ability to accurately replicate these modified words in the generated images. 

\paragraph{Random Word Combination (RWC)}
Our third strategy extends the concept of introducing complexity by randomly combining words to form new, potentially unseen phrases. Similar in intent to the second method, this approach is designed to further challenge the model's generalization capabilities. By presenting the model with these unconventional combinations, we aim to explore its limits in adhering to text prompts that may fall outside its pre-training data. To achieve this, we fist create a template ``A neon sign of \textit{placeholder}''. We also create a list of uncommon words, such as ``quickbonook", then we randomly choose words from this list combined with punctuations such as ``!, @, \%, \$, \&, \#, *, \^{}''.

\begin{table}[t]
\centering
 \resizebox{0.9\linewidth}{!}{
\begin{tabular}{c|c|c|c|c}
    \toprule
    \multirow{2}{*}{\textbf{Features}}& \multicolumn{3}{c|}{\textbf{LenCom-EVAL}}& \multirow{2}{*}{\textbf{MARIO-EVAL}}\\
     \cmidrule(lr){2-4}
     & {\textbf{MARIO-Hard}} &  {\textbf{Aug-MARIO-Hard}}  & {\textbf{RWC}} & \\
     \toprule
    Size & 1,000 & 1,000 & 1,000 & 4000\\
    Minimum Words & 4 & 4 & 1 & 1\\
    Maximum Words & 14 & 17 & 10 & 9\\
    Average Words & 4.9 & 5.8 & 5.5 & 2 \\
    \bottomrule
    \end{tabular}
    }
    \vspace{3mm}
\caption{Comparison of data statistics across our three subsets with the established benchmark MARIO-EVAL:  LenCom-EVAL facilitates the generation of visual text involving lengthy words usage.}
\label{tab:statistic}
\end{table}

\section{Findings of Glyph-Controlled Image Generation}
Given that two-stage glphy-controlled image generation~\cite{chen2023textdiffuser,chen2024textdiffuser,yang2024glyphcontrol,tuo2023anytext}, which first generate a text layout and then create an image based on this layout, achieve top-tier performance, we investigate the limitations of such systems. 
In Section~\ref{sec:preliminary}, we provide an overview of a representative model from this family, TextDiffusor~\cite{chen2024textdiffuser}, and identify its major limitations in Section~\ref{sec:limitations_tf}. These findings motivate us to design training-free approaches to enhance the system's capabilities, which we will introduce in Section~\ref{sec:auto_correct}.
\subsection{Preliminary} 
\label{sec:preliminary}
\paragraph{Layout Generator.}
To create bounding boxes (represented by four coordinates) for each keyword within the provided text prompts. TextDiffuser~\cite{chen2024textdiffuser} utilizes a Transformer-based Encoder-Decoder as the layout generator, albeit with a few modifications. The Encoder processes the text prompt embeddings, which consist of four components:
\begin{equation}
Embedding (P) = CLIP(P)+Pos(P)+Key(P)+Width(P),
\end{equation}
where $CLIP(P)$ represents the embeddings of $P$ produced by the CLIP model, $Pos(P)$ denotes the position embedding, $Key(P)$ identifies the embedding that indicates whether a token is a keyword, and $Width(P)$ captures the embedding of the character count in each token of $P$.
The decoder takes the positional embeddings as the query to ensure that the $n^{th}$ query corresponds to the ${n^{th}}$ keyword in the prompt. 
In contrast, TextDiffuser-2~\cite{chen2023textdiffuser} employs a Large Language Model (LLM) to create more varied layouts. Once the bounding boxes are generated, character-level masks are produced using Python libraries such as Pillow\footnote{https://pypi.org/project/pillow}. 
Both the models are trained on the MARIO dataset with the OCR generated bounding boxes. 

\paragraph{Visual Text Image Generator.}
The goal of the image generator is to leverage the generated segmentation masks and the text prompt as conditions for the image generation process. TextDiffuser \cite{chen2024textdiffuser} utilizes a VAE \cite{kingma2013auto} to encode the image, and the encoder's output feature is given to a Diffusion Model \cite{ho2020denoising} to corrupt and denoise a new image feature. This feature is finally passed to the VAE decoder to decode the image. This diffusion model is trained with denoising and character-aware losses, where the latter is provided by a pretrained U-NET \cite{ronneberger2015u} model that maps the latent image features to character-level segmentation masks.


\subsection{Limitations} 
\label{sec:limitations_tf}
While TextDiffuser can generate more accurate text compared to standard diffusion models such as Stable Diffusion XL (SD-XL), it still makes obvious mistakes which may render it unsuitable for practical applications. Below, we outline three key limitations of TextDiffuser.

\begin{minipage}[b]{.99\textwidth}
\includegraphics[width=0.45\linewidth]{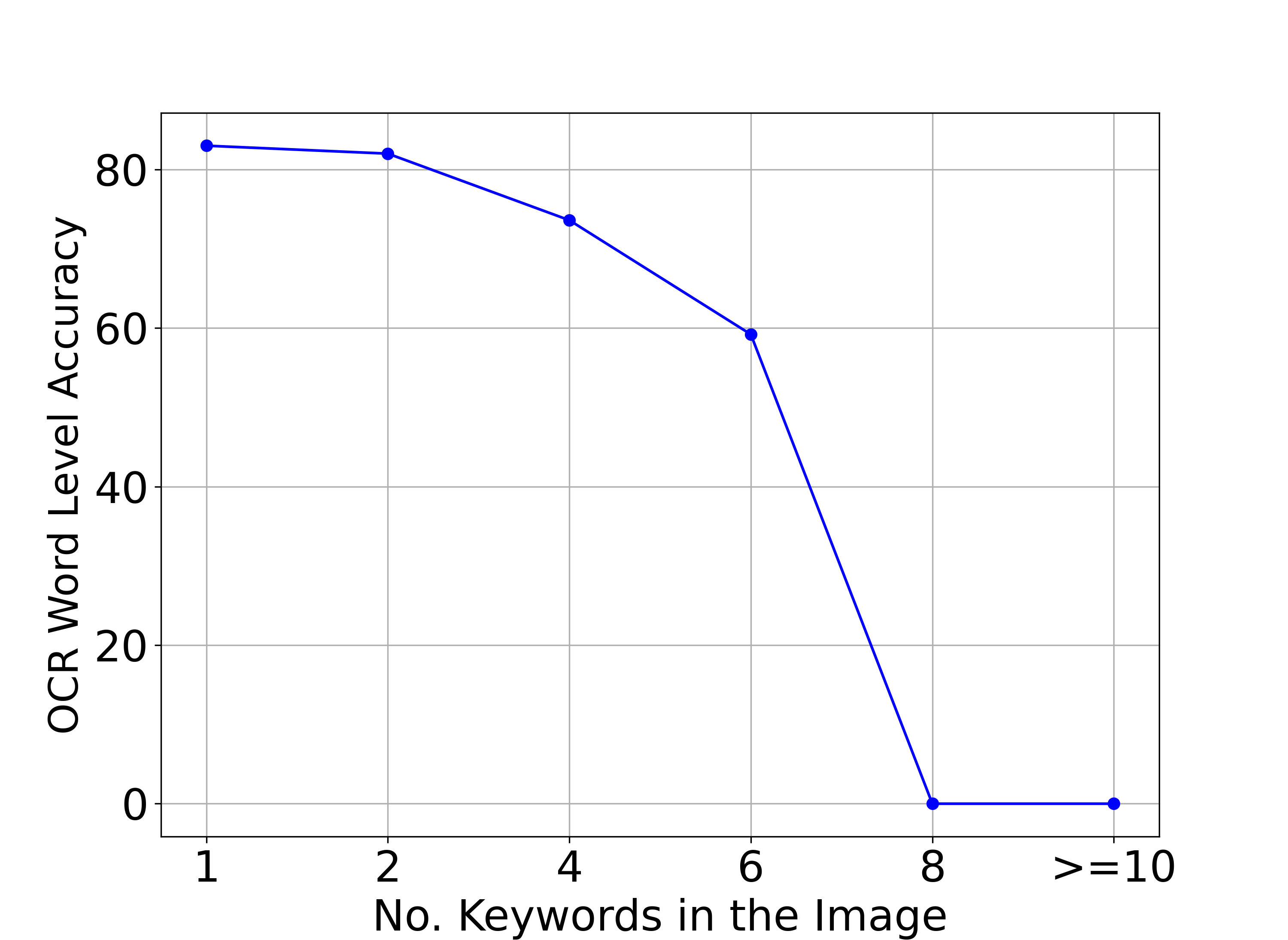}
\includegraphics[width=0.45\linewidth]{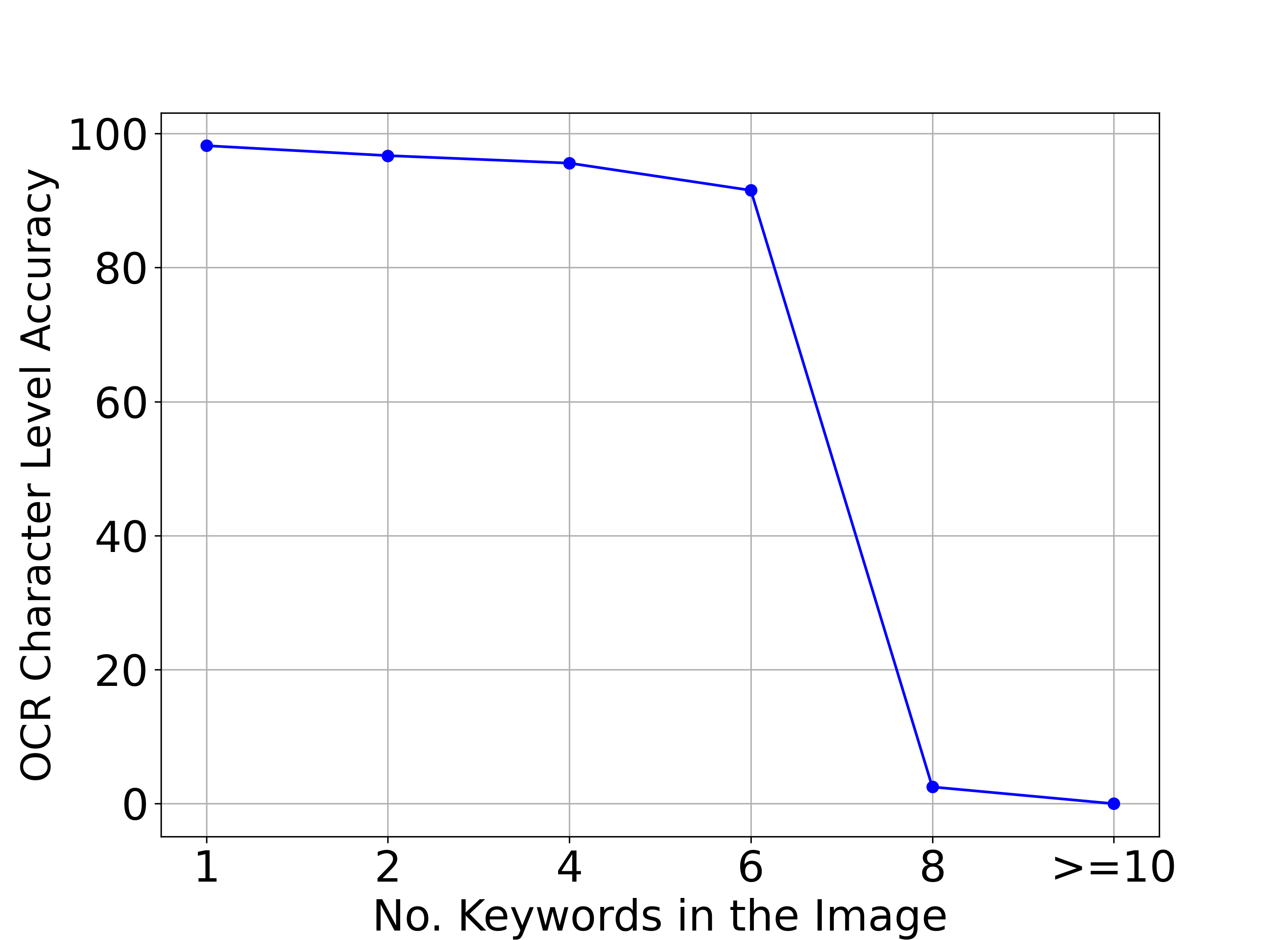}
\captionof{figure}{Accuracy of TextDiffuser across image subsets as the number of keywords increases: Performance drops as the number of words in the image increases.}
\label{fig:word-length-performance}
\end{minipage}%

\paragraph{Diminishing performance with increased text length.}
We found that TextDiffuser struggles with long text. Figure~\ref{fig:word-length-performance} demonstrates this by plotting the performance of TextDiffuser on 5 different subsets on MARIO-Hard dataset, with the number of keywords being varied for each subset in the text evaluated by OCR word-level F1 score (described in \S\ref{sec:eval_metrics}). 
As we can see from the figure that the performance drops while the length increase, and the performance is even closer to 0 when the number of words is more than 8 words.

\begin{figure}
    \centering
    \includegraphics[width=0.95\linewidth]{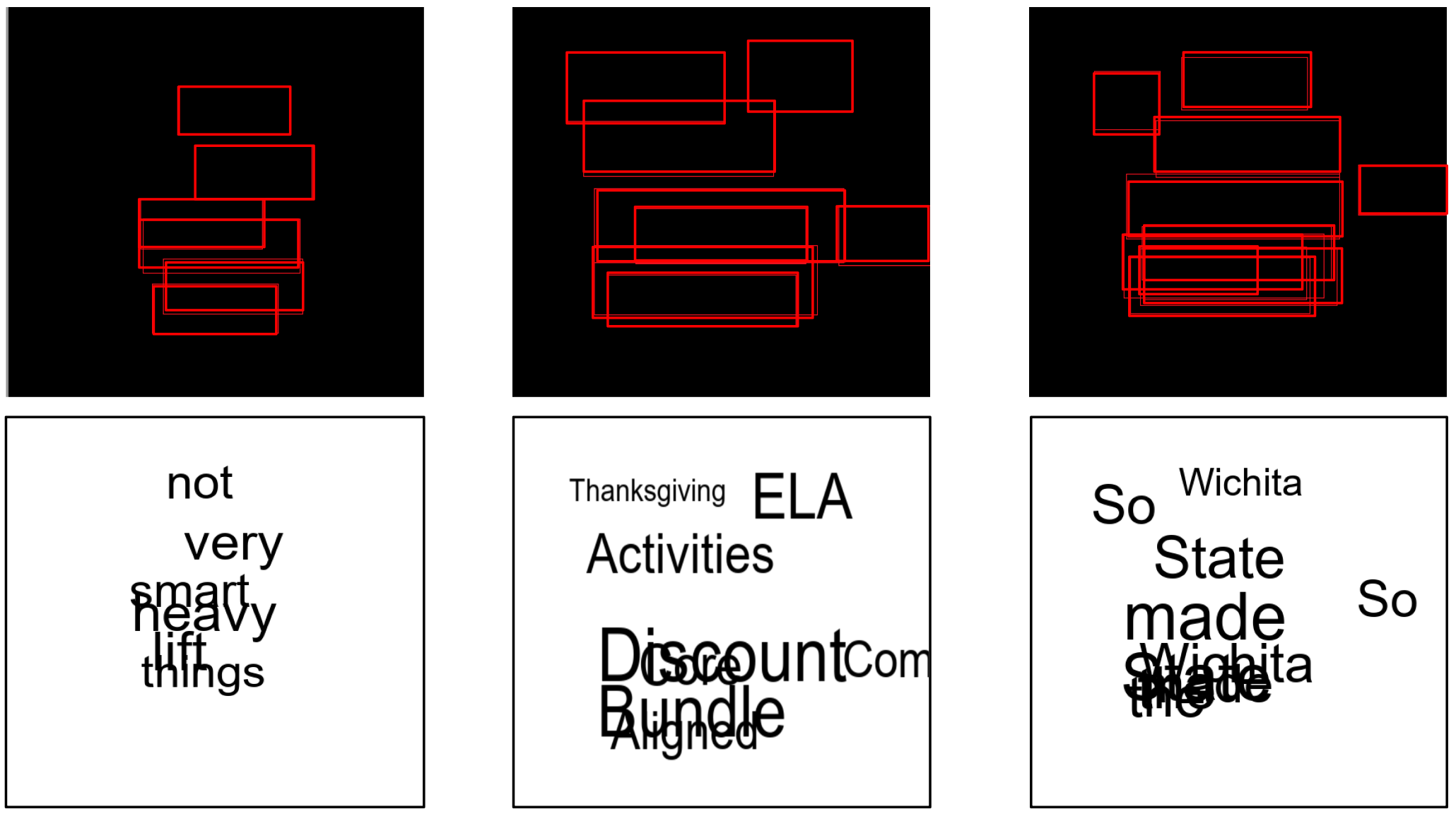}
    \caption{The generated bounding boxes and the corresponding glyph images with 6, 8, and 10 keywords: The overlapped area of the bounding boxes increases when the number of keywords increases.}
    \label{fig:layout}
\end{figure}

\paragraph{Poor layout generation leads to overlapping text.
}
In our experiments, we observed that the layout generator occasionally created bounding boxes for keywords that overlapped, leading to errors in the text visuals generated by the image generator, which depends on the layout's mask features.
Figure~\ref{fig:layout} displays the generated layouts for 6, 8, and 10 words, clearly showing an increase in the overlap area of the bounding boxes as the number of keywords increases.

\begin{figure}[h!]
    \centering
    \includegraphics[width=0.98\linewidth]{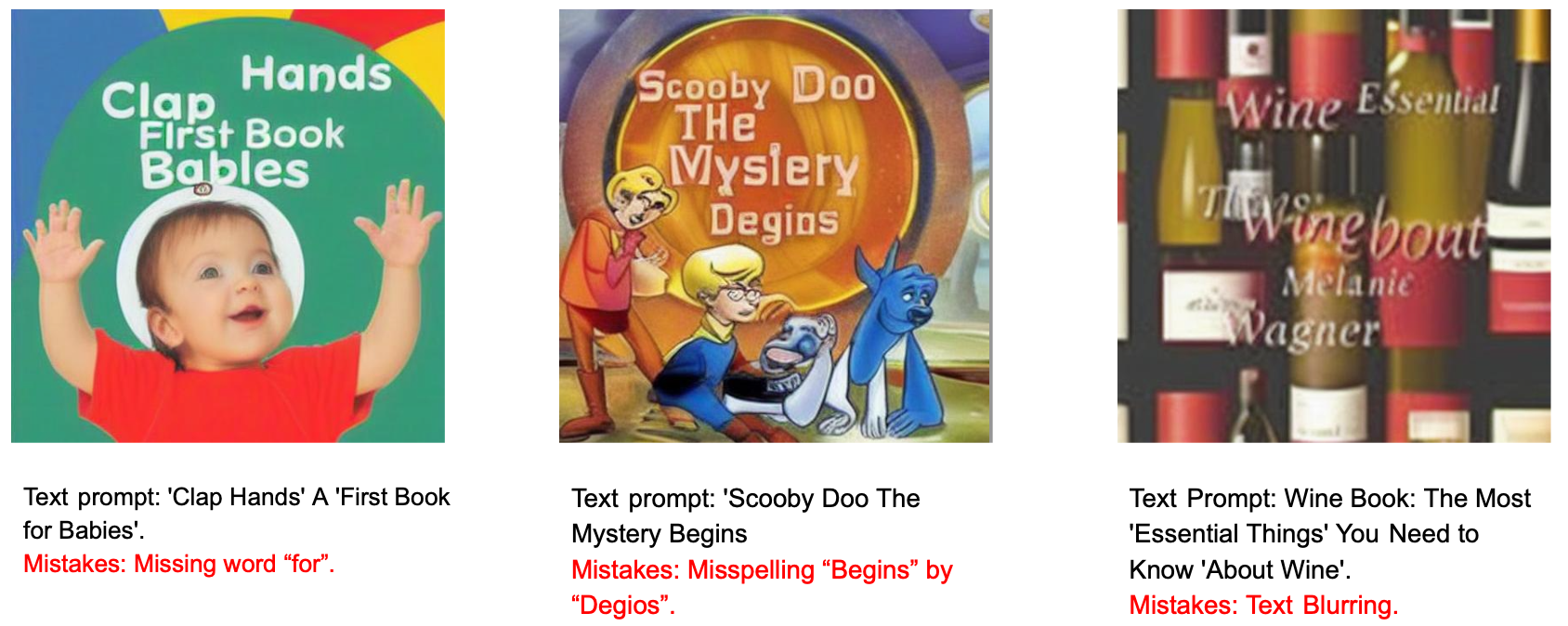}
    \caption{Examples of three frequent errors made by TextDiffuser: missing words (left), misspelling (middle), and blurry text (right).}
    \label{fig:misspelling}
\end{figure}

\paragraph{Inability to strictly adhere to text prompts.
}
Throughout our experiments, we discovered that TextDiffuser fails to generate  complete, accurate and clear textual representations within images in some cases. 
We identify three common mistakes as demonstrated in Figure~\ref{fig:misspelling}. 
\textit{ Missing words (Left)}: TextDiffuser fails to generate certain words from the original text prompt in the generated image
\textit{Misspellings (Middle)}: TextDiffuser fails to generate precise words.
\textit{Blurry Text (Right)}: TextDiffuser generates blurry text, making it difficult to read and understand.

\section{\textbf{S}imulated \textbf{A}nnealing and \textbf{OCR}-Aware Recursive In-\textbf{Paint}ing for Glyph-Enhanced Image Generation}
\label{sec:auto_correct}

Based on our pioneer study, minimizing bounding box overlap and correcting spelling errors are key factors in enhancing the accuracy of visual text. Therefore, we introduce two training-free methods to address these issues. 
These methods are designed to seamlessly integrate into any two-stage system, such as TextDiffuser, TextDiffuser-2, and AnyText. In this work, we  take TextDiffuser as our foundational system. 
In the following, we describe each method to reduce the overlap and mispelling in the generated images. 
\subsection{Simulated Annealing to Reduce the Overlap of Layout Generation} 

We employed Simulated Annealing (SA)~\cite{kirkpatrick1983optimization}, a gradient-free algorithm, to mitigate keyword overlap during the generation of glyph images. 
The general idea of SA is to reduce the \textit{energy} of the bounding boxes, which is measured by the total area of overlap of all bounding boxes. 
In real implementation, we found that assigning greater weights to larger bounding boxes proved to expedite convergence, enhancing efficiency. Consequently, we adopted a weighted overlap area as our energy metric. The procedure is detailed in the algorithm steps outlined.
Algorithm~\ref{alg_sa} outlines the overall steps. 
Initially, bounding boxes generated by the layout generator are input into the SA algorithm, and their energy is calculated (Line 2). A random adjustment function then modifies the arrangement of the bounding boxes (Line 3), leading to a new energy calculation (Line 4), denoted as \textit{energy'}. 
This adjustment introduces variability by altering the positions of the bounding boxes both horizontally and vertically.
we calculate the likelihood of accepting the new arrangement based on a probability function, with the probability influenced by a decreasing temperature parameter (Line 5). This temperature reduction strategy ensures a higher acceptance probability in early iterations, gradually making transitions less likely as the process progresses. Ideally, as the temperature approaches zero, the algorithm reaches the global minimum energy configuration. 
We initiated the process with a temperature value (modulated function) set at 1.0 and gradually decreased it at a cooling rate of $\frac{1}{3000}$ during each iteration. 
The maximum number of iterations was set to 70.

\begin{algorithm}
\caption{Simulated Annealing for Minimizing Keyword Bounding Boxes Overlap}\label{alg:cap}
\label{alg_sa}
\textbf{Input} Initial arrangements of keyword bounding boxes $\mathcal{BB}$, Initialized temperature $\mathcal{T}$ is 1.0, Cooling Rate $\mathcal{CR}$ is $\frac{1}{3000}$, the maximum iteration $max\_iter$ is 70, the initial iteration $iter$ is 0.
\textbf{Output} Reduced/No overlap bounding box rectangles $\mathcal{BB}$.
\begin{algorithmic}[1]
\While{Weighted\_Overlap($\mathcal{BB}$) and $iter < max\_iter$ }
    \State energy $\gets$ Weighted\_Overlap($\mathcal{BB}$)
    \State $\mathcal{BB'}$ $\gets$ Random\_Adjustment($\mathcal{BB}$) 
    \State energy' $\gets$ Weighted\_Overlap($\mathcal{BB'}$)
    \State $\mathcal{P}$ $\gets$ $e^{\frac{-(energy' - energy)}{\mathcal{T}}}$
    \If{random(0,1) < $\mathcal{P}$} 
        \State $\mathcal{BB} \gets \mathcal{BB'}$
    \EndIf
    \State $\mathcal{T} \gets \mathcal{T} - \mathcal{CR}$
\EndWhile
\end{algorithmic}
\end{algorithm}

\begin{figure}[h!]
    \centering
    \includegraphics[width=0.95\linewidth]{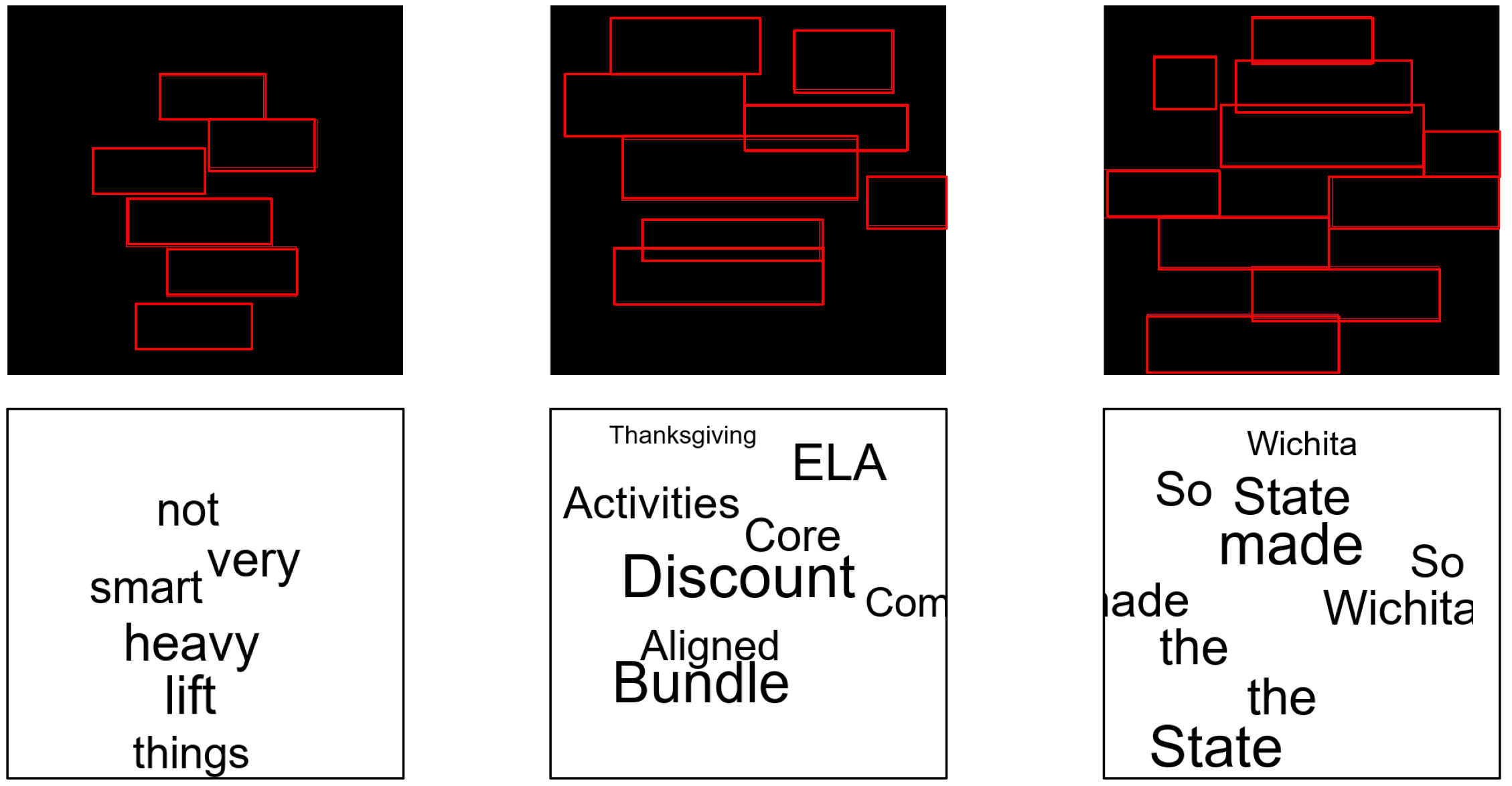}
    \caption{SA reduces the overlap of bounding boxes for keywords in the examples shown in Figure~\ref{fig:layout}.}
    \label{fig:correct_layout}
\end{figure}

\paragraph{Visualization of Layout Correction.}
We apply the SA algorithm to the examples shown  in Figure~\ref{fig:layout}. As demonstrated in Figure~\ref{fig:correct_layout}, after searching for a new arrangement, the final layout exhibits minimal overlap, and the clarity of the generated glyph images is significantly improved. 

\subsection{OCR-Aware Recursive In-painting to Mitigate Misspellings of the Image Generator} 
The images generated after the diffusion process are prone to misspelling, resulting in either missing characters or the presence of unnecessary characters in words. 
To address this challenge, we propose a recursive spelling correction algorithm for the output image. We utilize the original generated image as a reference and detect misspelled words using Paddle OCRv3~\cite{li2022pp}. Then we compare the extracted text with the ground truth word-by-word, and generate a glyph figure with the misspelling word regions but with the correct spelling words. 
Subsequently, this new glyph figure along with the original image and the text prompt are given to an in-painting model. 
Since the image generation model in TextDiffuser can also be an image in-painting model, we use it for the misspelling correction task. 
This process is repeated for 2 iterations to effectively address the issue of misspelling. During our experiments, we found that 2 iterations yields the best performance while increasing the iterations decrease the quality of generating images. 

\begin{figure}[t]
    \centering
    \includegraphics[width=0.95\linewidth]{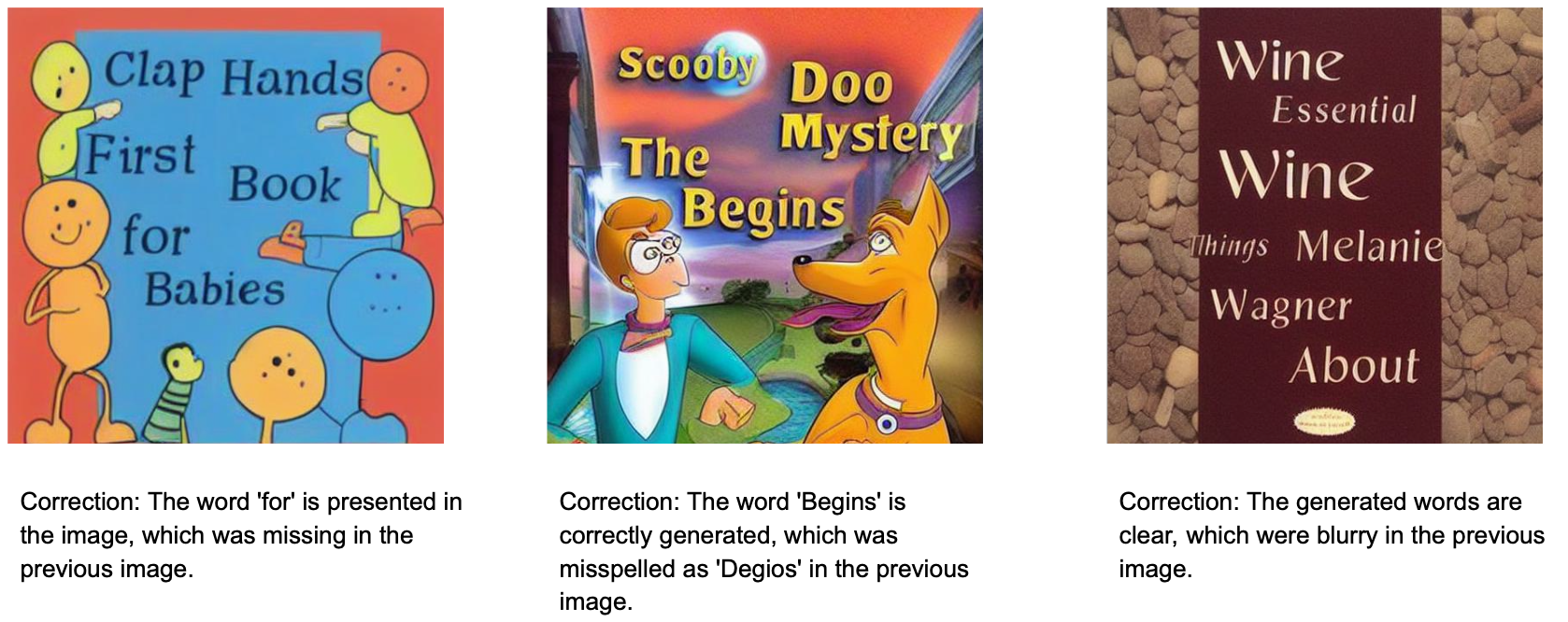}
    \caption{The examples generated by SA-OcrPaint and TextDiffuser given the same text prompts in the Figure~\ref{fig:misspelling}: SA-OcrPaint addresses and corrects prior errors, yielding text images that are notably more accurate and legible.}
    \label{fig:correct_spelling}
\end{figure}

\paragraph{Visualization of Misspelling Correction.}
Together with the SA and OCR recursive in-painting, we integrate SA-OcrPaint with TextDiffuser and using the same prompts given in Figure~\ref{fig:misspelling} to generate images. As shown by Figure~\ref{fig:correct_spelling}, SA-OcrPaint enhances the visual text output without missing words, misspelling and burry text.

\section{Experiments}
In this section, we first introduce all the baselines and any adjustments made to them for the experimental setup to ensure a fair comparison. 
For the implementation of SA, we adopted the codebase\footnote{https://github.com/mwkling/rectangle-overlap} and incorperated our weighted overlap adjustment.
Then, we describe the evaluation metrics used to measure performance. And lastly, we present the quantitative results. 
\subsection{Baselines}
\paragraph{TextDiffuser:} 
Similar to original Text-Diffuser \cite{chen2024textdiffuser} we used the 2 stage network with maximum keywords length of 8. For the diffusion process we used the provided pretrained check point of \texttt{runwayml/stable-diffusion-v1-5} and utilized Hugging Face Diffuser API\footnote{https://huggingface.co/docs/diffusers}. For inference, we used 30 sampling steps and classifier free guidance with the scale of 7.5.  We used a single NVIDIA RTX A6000 GPU with 48GB VRAM. Inference across all the datasets we experimented with was completed within 3 hours.
 
\paragraph{TextDiffuser-2:} The TextDiffuser-2~\cite{chen2023textdiffuser} pipeline consists of two language models: one for converting the input prompt into a language-format layout and the other for encoding this layout within the diffusion model to generate images. As described in the experimental setup, we sample with 50 steps during the inference phase of the pipeline. We use the pretrained \texttt{vicuna-7b-v1.5} model checkpoint of \texttt{JingyeChen22/textdiffuser2\_layout\_planner} for layout planning. For the diffusion process, we use the pre-trained checkpoint of \texttt{JingyeChen22/textdiffuser2-full-ft}.


\paragraph{Diff-Text:} For the Diff-Text~\cite{zhang2023brush} model, we use the official GitHub implementation along with the following pre-trained models: \texttt{runwayml/stable-diffusion-v1-5} and \texttt{lllyasviel/sd-controlnet-canny}. Furthermore, since the sketch renderer does not support line breaks, we truncate any text that goes beyond the 512 pixel limit. 

\paragraph{AnyText:} The AnyText~\cite{tuo2023anytext} inference pipeline contains a text-control diffusion pipeline with two primary components: an auxiliary latent module and a text embedding module. We use the official implementation on GitHub and use the \texttt{text-generation} mode. Following the experimental setup described in AnyText, we split each input into a maximum of 5 text lines, with each line having no more than 20 characters. Any text exceeding 100 characters was truncated. This input was then sent to the Glyph Builder, and the produced templates were fed into the pipeline to generate the final images with the inscribed text.

\subsection{Evaluation Metric} 
\label{sec:eval_metrics}
Our evaluation of textual accuracy relies on three primary metrics. The first one is CLIPScore~\cite{hessel2021clipscore}, which assesses the cosine similarity between the image and text representations derived from the CLIP model~\cite{radford2021learning}. 
The second metric is OCR (Optical Character Recognition) evaluation, for which we employ PaddleOCR\footnote{https://pypi.org/project/paddleocr} to recognize text within the generated images, and then we compute Precision, Recall, F1, Accuracy scores at both the word and character levels with the ground truth.
The last one is the Normalized Levenshtein Distance (NLD) \cite{Yujian2007nld} which measures the edit distance between two strings based on the minimum cost required to transform one string into the other via a series of weighted editing operations.
Although FID~\cite{heusel2017gans} have been used in previous studies, it is unsuitable for assessing visual text accuracy, as evidenced by \cite{chen2023textdiffuser} that FID Score is not consistent with other metrics.

\subsection{Results}
\subsubsection{LenCom-EVAL.}

\begin{table*}[t]
\small
\centering
\setlength\tabcolsep{5pt}
\resizebox{.98\textwidth}{!}{
\begin{tabular}{clccccccccccc}
\toprule
\multirow{2}{*}{\textbf{Dataset}}  & \multirow{2}{*}{\textbf{Model}} &  \multirow{2}{*}{\textbf{CLIPScore~$\uparrow$}} & \multicolumn{4}{c}{\textbf{OCR Character-Level~$\uparrow$}}  & \multicolumn{4}{c}{\textbf{OCR Word-Level~$\uparrow$}} & \multirow{2}{*}{\textbf{NLD~$\downarrow$}}\\
\cmidrule(lr){4-7}  \cmidrule(lr){8-11} 
~ & ~ & ~ & Pre & Rec & F1 & Acc & Pre & Rec & F1 & Acc   \\
\midrule
\multirow{6}{*}{\rotatebox[origin=c]{0}{MH}} 
&TextDiffuser & 36.74 & 0.92 & 0.83 & 0.87 & 0.89 & 0.64 & 0.69 & 0.64 & 0.66 & 27.12 \\
&TextDiffuser-2 & 33.41 & 0.86 & 0.75 & 0.80 & 0.85 & 0.63 & 0.58 & 0.60 & 0.58 & 26.70 \\
&AnyText  & 15.08 & 0.74 & 0.62 & 0.67 & 0.73 & 0.17 & 0.18 & 0.18 & 0.18 & 63.28 \\
&Diff-Text & 18.18 & 0.52 & 0.43 & 0.47 & 0.53 & 0.04 & 0.03 & 0.04 & 0.04 & 85.35 \\ 
& SA (Ours)  & {37.44} & {0.96} & {0.94} & {0.95} & {0.97} & {0.79} & {0.78} & {0.78} & {0.79} & {19.92}\\ 
& SA-OcrPaint (Ours)  & \textbf{37.56} & \textbf{0.97} & \textbf{0.95} &\textbf{ 0.96} & \textbf{0.98} & \textbf{0.86} & \textbf{0.88} &\textbf{ 0.87} & \textbf{0.88} & \textbf{18.55} \\ 

\midrule 
\multirow{6}{*}{\rotatebox[origin=c]{0}{Aug-MH}} 
 & TextDiffuser & 35.02 & 0.80 & 0.65 & 0.72 & 0.76 & 0.57 & 0.51 & 0.53 & 0.51 & 43.70 \\
 & TextDiffuser-2 & 30.47 & 0.84 & 0.63 & 0.73 & 0.80 & 0.43 & 0.34 & 0.38 & 0.34 & \textbf{24.42} \\
 & AnyText & 15.32 & 0.74 & 0.57 & 0.64 & 0.73 & 0.14 & 0.13 & 0.14 & 0.13 & 66.38  \\
 & Diff-Text & 20.09 & 0.56 & 0.35 & 0.43 & 0.49 & 0.01 & 0.009 & 0.01 & 0.009 & 85.25 \\ 
 & SA(Ours) & {37.28} & {0.96} & {0.87} & {0.91} & {0.96} & {0.77} & {0.74} & {0.75} & {0.74} & {29.95} \\ 
 & SA-OcrPaint (Ours) & \textbf{37.28} & \textbf{0.97} & \textbf{0.89} & \textbf{0.93} & \textbf{0.97} & \textbf{0.84} & \textbf{0.83} & \textbf{0.83} & \textbf{0.83} & \textbf{28.47} \\ 

\midrule 
\multirow{6}{*}{\rotatebox[origin=c]{0}{RWC}} 
& TextDiffuser  & 37.96 & 0.88 & 0.76 & 0.82 & 0.81 & 0.62 & 0.52 & 0.57 & 0.53 & 32.51 \\
& TextDiffuser-2 & 33.22 & 0.95 & 0.79 & 0.86 & 0.86 & 0.68 & 0.57 & 0.62 & 0.57 & 26.76 \\
& AnyText  & 23.67 & 0.69 & 0.51 & 0.59 & 0.61 & 0.12 & 0.15 & 0.13 & 0.15 & 69.56\\
& Diff-Text & 19.33 & 0.50 & 0.48 & 0.49 & 0.56 & 0.03 & 0.02 & 0.02 & 0.02 & 85.63 \\ 
& SA (Ours) & 39.23 & 0.95 &  0.88 & 0.92 & 0.91 & 0.72 & 0.66 & 0.68 & 0.66 & 25.96 \\ 
& SA-OcrPaint (Ours)  & \textbf{39.23} & \textbf{0.97} & \textbf{0.91} & \textbf{0.94} & \textbf{0.92} & \textbf{0.80} & \textbf{0.74} & \textbf{0.77} & \textbf{0.74} & \textbf{23.68} \\  
\bottomrule
\end{tabular}
}
    \vspace{3mm}
\caption{Comparison of performance of our method with baseline and other existing models across three subsets (MH: MARIO-Hard): our approach significantly outperforms the baseline, achieving the best performance. }
\label{table:main_result}
\end{table*}


We conduct two experiments based on our proposed method. One is to just apply SA algorithm to fix the layout generation (referred as SA), and the second one is to apply both SA and OCR in-painting (referred as SA-OcrPaint). 
Table~\ref{table:main_result} shows that both methods achieve significant gains compared the existing methods across all the metrics on the three subsets. Noticeably, compare to the TextDiffuser, which is our backbone model, our framework increase the performance significantly. 
For instance, compare to TextDiffuser, SA enhance the OCR F1 on the word-level by 14\%, 22\%, and 11\% on MARIO-HARD, Aug-MARIO-HARD, and RWC respectively. 
The effectiveness of SA is because of the large reduction of the overlapping area of keyword bounding boxes, and more quantitative results are represented in Appendix~\ref{apd:overlap}. 
SA-OcrPaint make even more improvements: 23\%, 30\%, and 20\% on the three subsets respectively. These demonstrates the effectiveness of the proposed SA and OCR aware in-painting technique. 
We further evaluate the accuracy on different keyword length. We observe that the improvement is more significant when the keyword length increase. More results can be found in Appendix~\ref{apd:length}.  
As we mentioned previously, our framework can plug-in any two-stage systems which first generate the keywords layout and then the visual text images, we expect that our framework can also improve TextDiffuser-2 and AnyText systems. We leave this as future investigation. 
By comparing the performance across different subsets, it is easy to see that Aug-MARIO-HARD and RWC are harder than MARIO-HARD, which is align with our hypothesis that the existing systems are struggle with rare words generation. This implies that rare words generation is another challenges for diffusion models.  

\subsubsection{MARIO-EVAL.}

Here, we also evaluate the propose method SA-OcrPaint method on the MARIO-EVAL datasets.
In Table~\ref{tab:mario-eval}, we compare SA-OcrPaint with the previous systems. Note that the accuracy metric in the original paper~\cite{chen2023textdiffuser} is the exact match in the sentence level (referred as OCR Accuracy [Sent]), for the sake of fair comparison, we also compute this at the sentence level rather than character or word level like Table~\ref{table:main_result}.
It is easy to see that our SA-OcrPaint outperforms exisiting models by large margin and achieve the state-of-the-art performance on MARIO-EVAL benchmark. 

\begin{table}[ht]
\centering
\resizebox{0.98\textwidth}{!}{%
\begin{tabular}{ccccccc}
\toprule
 \textbf{Metrics} & \textbf{SD-XL} & \textbf{PixArt-$\alpha$ } & \textbf{GlyphControl} & \textbf{TextDiffuser} & \textbf{TextDiffuser-2} & \textbf{SA-OcrPaint}\\
\toprule
CLIPScore  & 31.31 & 27.88 & 34.56 & 34.36 & 34.50  & \textbf{35.70}\\
OCR-Word F1  & 3.66 & 0.03 & 64.07 & 78.24 & 75.06 & \textbf{85.70} \\
OCR Accuracy [Sent] & 0.31 & 0.02 & 32.56 & 56.09 & 57.58 & \textbf{66.95}\\

\bottomrule
\end{tabular}%
}
\caption{Comparison with other models on MARIO-EVAL. All the numbers are gathered from TextDiffuser-2 paper, and SA-OcrPaint achieves the new state-of-the-art results on MARIO-EVAL.}
\label{tab:mario-eval}
\end{table}

\subsection{Qualitative Examples}

\begin{figure}
    \centering
    \includegraphics[width=0.98\linewidth]{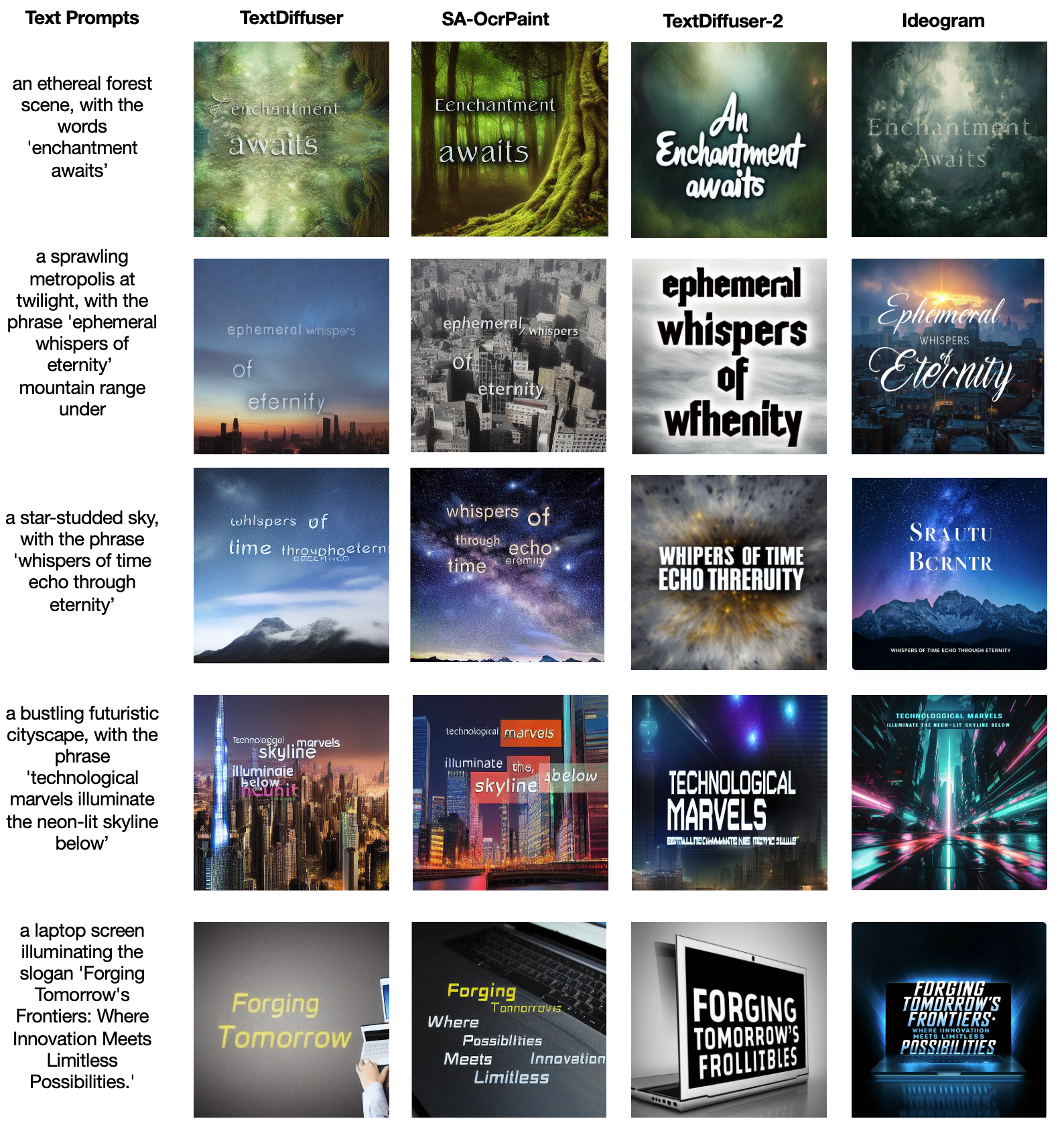}
    \caption{Examples of generated images from different models.}
    \label{fig:multiple-modal}
\end{figure}

To show qualitative analysis, we crafted text prompts by hand instead of relying on any established evaluation benchmarks to compare images generated by various models: TextDiffuser, TextDiffuser-2, SA-OcrPaint, and Ideogram, where the last model is only available through API\footnote{https://ideogram.ai}. 
In the comparative analysis illustrated in Figure~\ref{fig:multiple-modal}, several models' performances on image generation with textual content were evaluated. In the first prompt, all models accurately generated the text, with SA-OcrPaint producing noticeably clearer text than its counterparts. 
For the second prompt, TextDiffuser rendered the word ``eternity'' as the unclear ``efernify'', and TextDiffuser-2 introduced a spelling mistake, while SA-OcrPaint's output remained unambiguous. 
The third prompt shows TextDiffuser amalgamating words together, TextDiffuser-2 omitting the final word, and Ideogram adding extra, irrelevant text in the middle of the image. In contrast, SA-OcrPaint alone accurately rendered the intended text. 
With the fourth prompt, TextDiffuser's output overlapping keywords, TextDiffuser-2 accurately generated only the initial two words with subsequent ones becoming indistinct. SA-OcrPaint consistently outperformed the others in clarity and accuracy of text generation across these tests.
The final prompt highlighted a unique challenge: TextDiffuser and TextDiffuser-2 partially generated keywords, whereas SA-OcrPaint managed to produce all intended keywords, albeit misplacing them outside the laptop screen, indicating an area for future improvement in aligning text location with the prompt specifics.
Overall, SA-OcrPaint demonstrated superior capability in generating visually accurate textual content within images, outshining other models in comparison. This underscores its effectiveness, while also highlighting the importance of further refining text placement to match prompt instructions as a key direction for future research.






\section{Discussion and Conclusion}

In this study, we investigate on the generation of visual text images, a critical area within real-world image generation applications. 
Initially, we identified that while existing models like TextDiffuser can accurately generate text to a degree, they falter with lengthy and uncommon text. To advance research in this domain, we developed a testbed specifically aimed at generating long and rare visual text. 
Then we propose a training-free framework to enhance the performance of visual text generation model. 
Our approach was inspired by the observation that TextDiffuser often produces overlapping bounding boxes and spelling errors in its output. To minimize bounding box overlap, we implemented a simulated annealing algorithm with an adapted weighted energy function for more rapid convergence. 
Furthermore, to address spelling inaccuracies, we introduced an OCR-Aware Recursive In-painting method to correct errors in initially generated images. Our experimental results confirm the effectiveness of our proposed methods across all three subsets of our benchmark, demonstrating significant improvements in visual text image generation.

\noindent\textit{Limitation} 
Although a portion of our dataset originates from real-world situations, there remains a discrepancy due to the data augmentation and the use of random word combinations. A dataset enriched with more real-life scenarios would be substantially beneficial. Furthermore, we observed that during the recursive in-painting iterations, even after two iterations, the image generation model can still make spelling mistakes.

\clearpage  

%
%
\bibliographystyle{splncs04}
\bibliography{man}
\clearpage
\appendix 
\section{Dataset Details}

\subsection{Distribution }
\label{apd:dataset_distribution}
In this analysis, we present the distributions of ground truth word lengths within our LenCom-EVAL and MARIO-EVAL datasets. Figure \ref{fig:distribution} shows that LenCom-EVAL exhibits a higher proportion of lengthy keywords in its composition. 

\begin{minipage}[b]{.95\linewidth}
\includegraphics[width=0.49\linewidth]{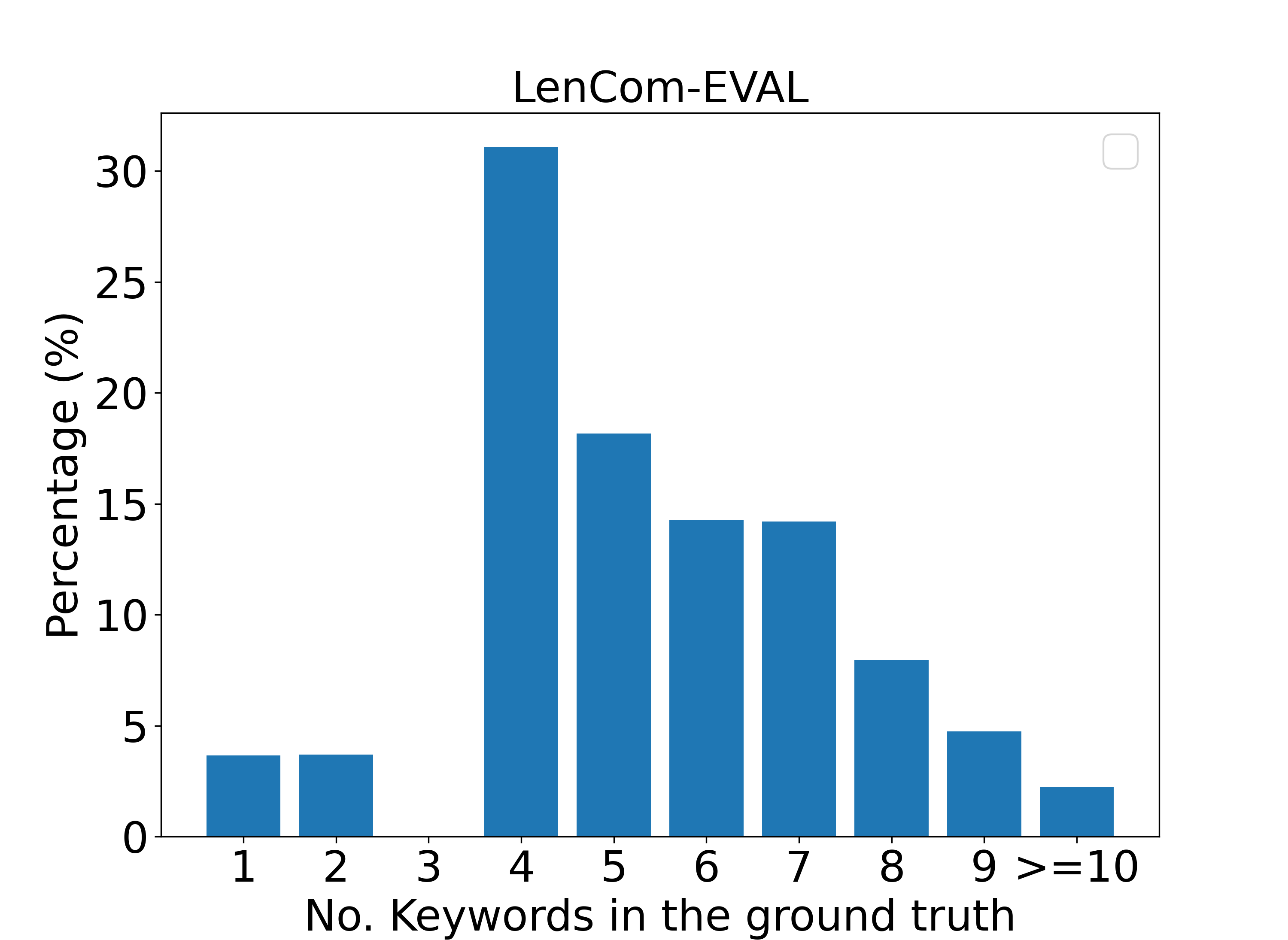}
\includegraphics[width=0.49\linewidth]{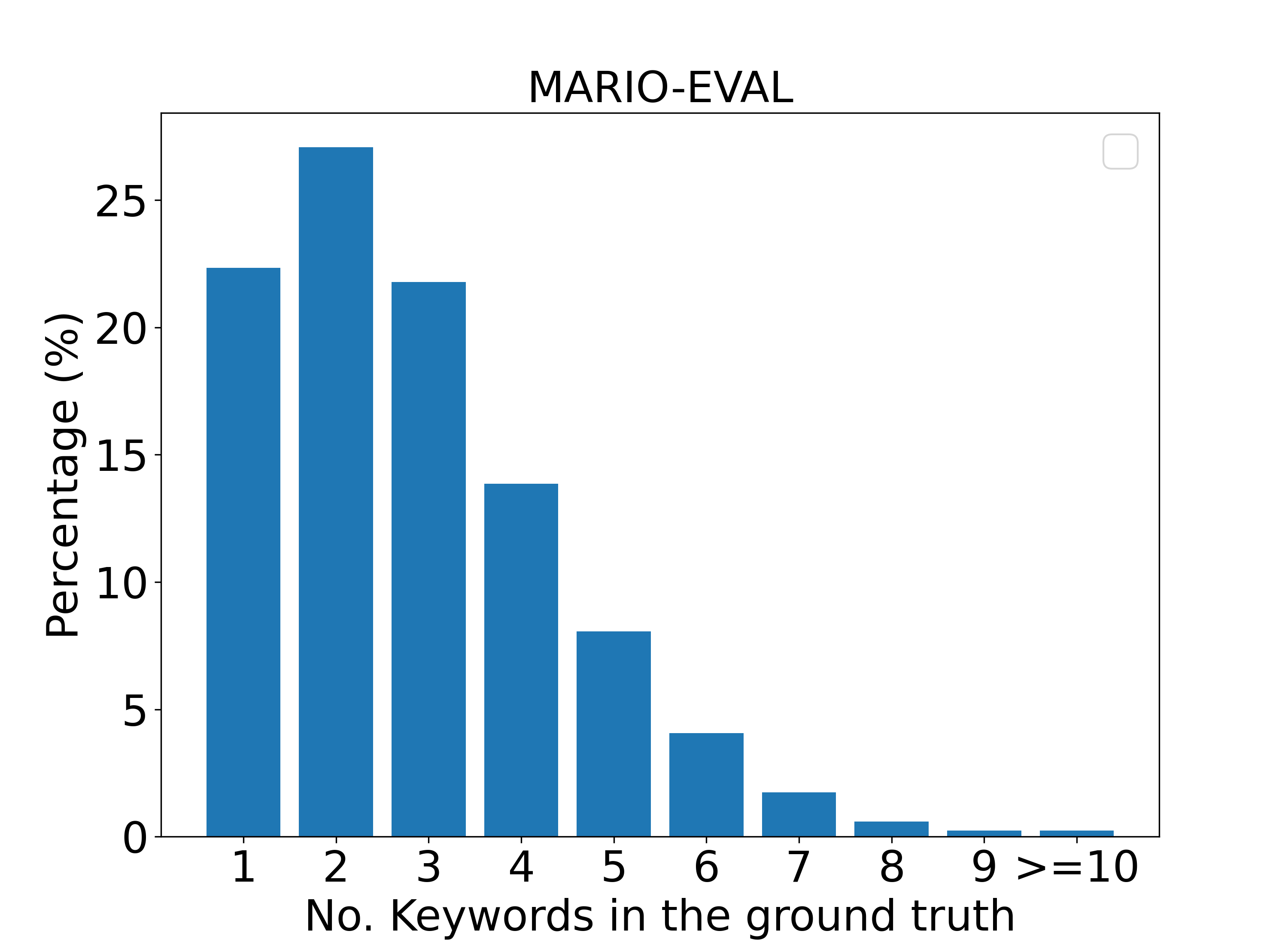}
\captionof{figure}{Comparison of LenCom-EVAL and MARIO-EVAL Distributions across Image Subsets by Number of Keywords: LenCom-EVAL exhibits a higher proportion of lengthy keywords in its composition. }
\label{fig:distribution}
\end{minipage}%

\subsection{Examples from Datasets}
\label{apd:dataset_examples}

Table~\ref{tab:examples_LenCom} displays an example from each subset within the LenCom-EVAL benchmark. In the Aug-MARIO-Hard instance, we employed the splitting augmentation method to divide the word "Amazon" into two parts: ``Amaz" and ``on". Our RWC subset is generated using the template "A neon sign that says [Placeholder]"; in this particular case, the placeholder was substituted with "The little Luminous in the garden".

\begin{table*}[h]
    \centering 
    \small
       \setlength{\tabcolsep}{4pt}
    \begin{tabular}{@{}p{0.1\linewidth} p{0.5\linewidth} p{0.3\linewidth} }
        \toprule
        \textbf{Subset} & \textbf{Input Text Prompt} & \textbf{Ground Truth} \\
        \toprule
        MARIO-Hard & `Amazon Cloud Player Amazon Cloud Player' Music & Amazon Cloud Player Amazon Cloud Player\\
        \hline
        Aug-MARIO-Hard & `Amaz on Clo ud Player Amazon Cloud Pla yer' Music & Amaz on Clo ud Player Amazon Cloud Pla yer \\
        \hline
        WRC & A neon sign that says `The little Luminous in the garden' & The little Luminous in the garden \\ 
    \bottomrule
    \end{tabular}
    \caption{
    Example from each subset in LenCom, the input prompt and the ground truth. 
    }
    \label{tab:examples_LenCom}
\end{table*}

\section{Experiments}

\subsection{Results for the Overlapped Area of Layout Generation}
\label{apd:overlap}

In addition to accuracy, we evaluate the overlap area and Intersection over Union (IoU) of bounding boxes generated by the layout generator. A smaller overlapped area or IoU indicates better layout generation by the model. From Table \ref{tab:iou}, we observe that SA-OcrPaint achieves significant reductions in both metrics. Specifically, our method improves the overlapped area and IoU by 91.3

\begin{table}[ht]
\centering
\begin{tabular}{ccc}
\toprule
 \textbf{Model}& \textbf{Overlapped Area} & \textbf{IoU}\\
\toprule
TextDiffuser  & 1450 & 0.41 \\
\hline
SA-OcrPaint  & 189 & 0.04 \\
\bottomrule
\end{tabular}
\caption{Comparison of TextDiffuser and SA-OcrPaint in terms of Overlapped Area and IoU: SA-OcrPaint demonstrates a notable reduction in the extent of overlapped bounding boxes. }
\label{tab:iou}
\end{table}

\subsection{Accuracy on 1/2/4/6/8/10 words subsets of SA-OcrPaint}
\label{apd:length} 

We present the OCR accuracy at both the word and character levels across various subsets. From Figure~\ref{fig:word-length-performance-2}, it's evident that SA-OcrPaint outperforms the baseline TextDiffuser notably when there are two or more keywords, with the improvement becoming more significant as the number of words increases. Conversely, when there's only one keyword, TextDiffuser performs better. In this scenario, where there's no overlap of bounding boxes due to a single bounding box in the layout generation, our SA algorithm doesn't exert any influence. However, during the second OCR in-painting step, multiple iterations actually decrease model performance. A potential remedy for this issue is to introduce a policy for accepting either the newly generated image or the previous one. This policy could be based on OCR word or character accuracy, wherein if the accuracy of the new image surpasses that of the previous one, it is accepted; otherwise, it is rejected.

\begin{minipage}[b]{.95\linewidth}
\includegraphics[width=0.49\linewidth]{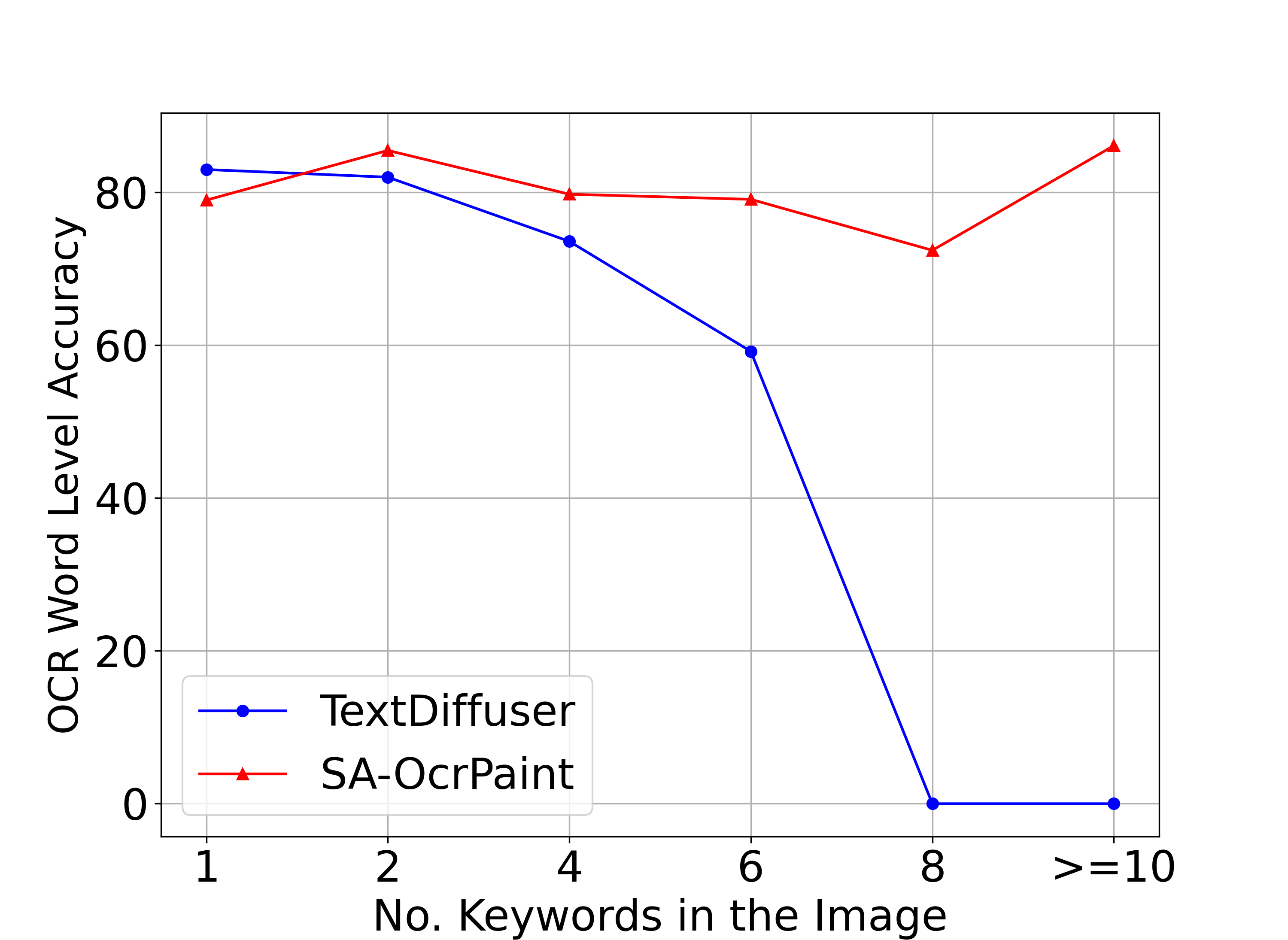}
\includegraphics[width=0.49\linewidth]{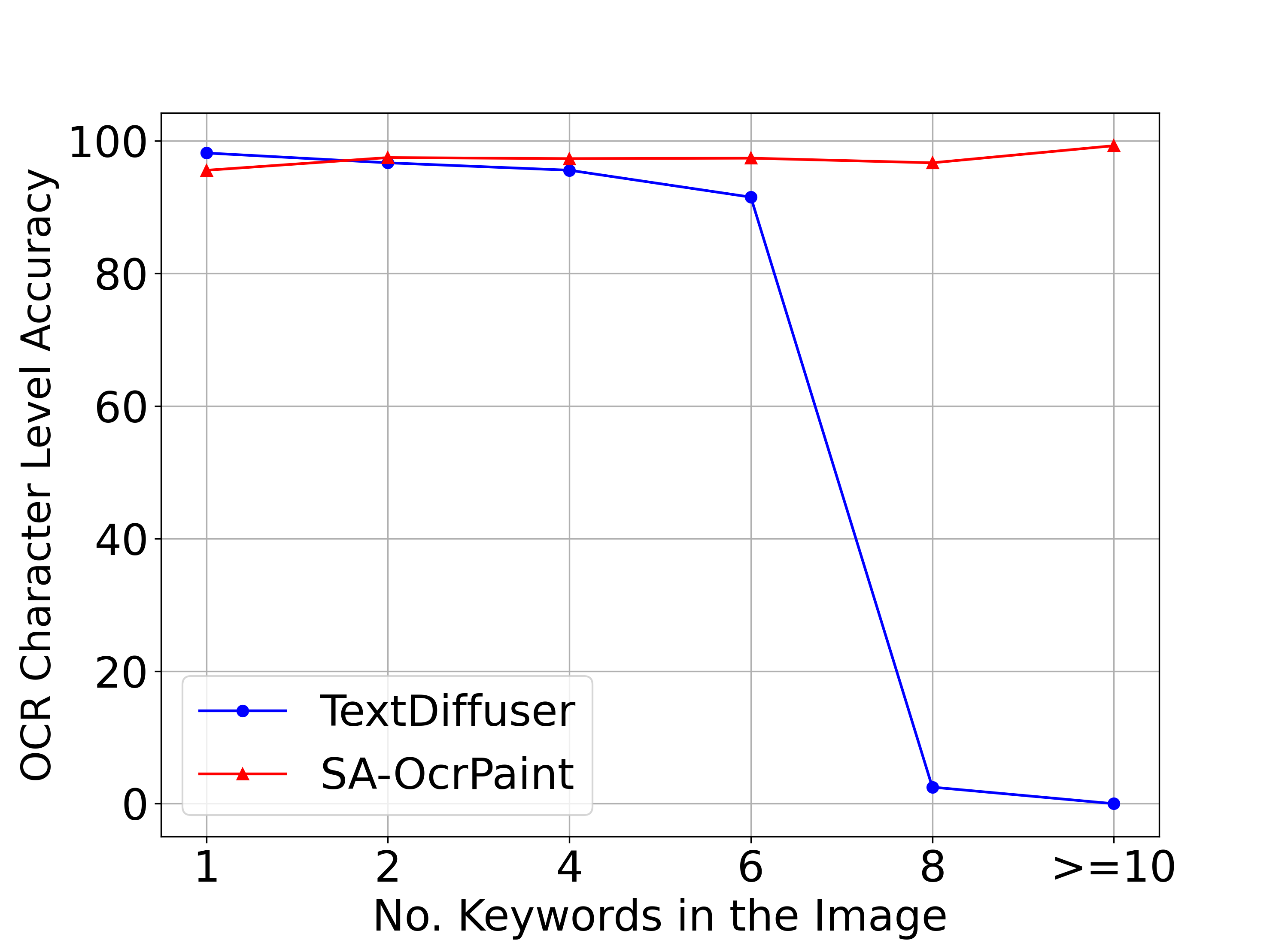}
\captionof{figure}{Compare Accuracy of TextDiffuser and SA-OcrPaint across image subsets as the number of keywords increases: SA-OcrPaint outperforms TextDiffuser when keywords number is euqal or more than 2.}
\label{fig:word-length-performance-2}
\end{minipage}%

\end{document}